\ifdefined\pdfsuppressptexinfo\pdfsuppressptexinfo=-1\fi
\ifdefined\pdfminorversion\pdfminorversion=7\fi
\documentclass[conference]{IEEEtran}
\IEEEoverridecommandlockouts    

\usepackage{cite}
\usepackage{multirow}
\usepackage{amsmath}
\usepackage{amssymb}
\usepackage{bm}
\usepackage{booktabs}
\usepackage[ruled,vlined,linesnumbered]{algorithm2e}
\usepackage{graphicx}
\usepackage[caption=false,font=footnotesize]{subfig}
\usepackage{xcolor}
\usepackage[hidelinks]{hyperref}
\graphicspath{{./figs/}}

\newcommand{\vx}{\mathbf{x}}
\newcommand{\vy}{\mathbf{y}}
\newcommand{\vz}{\mathbf{z}}
\newcommand{\hvy}{\hat{\mathbf{y}}}
\newcommand{\zS}{z_{S}}
\newcommand{\vzI}{\mathbf{z}_{I}}
\newcommand{\Lpref}{\mathcal{L}_{\text{pref}}}
\newcommand{\Lbase}{\mathcal{L}_{\text{base}}}
\newcommand{\Phat}{\hat{P}}
\newcommand{\minADE}{\text{minADE}_5}
\newcommand{\minFDE}{\text{minFDE}_5}

\title{\LARGE \bf Controllable Generative Trajectory Prediction\\ via Weak Preference Alignment}

\newif\ifanonymous

\ifanonymous
\author{\parbox{\textwidth}{\centering Anonymous Authors}}
\else
\author{
	\parbox{\textwidth}{%
		\centering
		Yongxi Cao$^{1,2,*}$ \thanks{* Work done during the Master's thesis at TU Delft.}, Julian F. Schumann$^{1}$, Jens Kober$^{3}$, Joni Pajarinen$^{2}$, Arkady Zgonnikov$^{1}$%
	}%
	\thanks{$^{1}$Delft University of Technology, Delft, the Netherlands {\tt\small \{j.f.schumann, a.zgonnikov\}@tudelft.nl}, $^{2}$Aalto University, Espoo, Finland	{\tt\small \{yongxi.cao, joni.pajarinen\}@aalto.fi}, $^{3}$University of Stuttgart, Stuttgart, Germany \tt\small jens.kober@ki.uni-stuttgart.de}%
}
\fi

\begin{document}
	\bstctlcite{IEEEexample:BSTcontrol} 

	\maketitle
	\thispagestyle{empty}
	\pagestyle{empty}

\begin{abstract}
Planners for automated vehicles rely on predictions of how surrounding road users may behave. Generative predictors capture multimodal futures well, but their samples are governed by latent variables with no semantic meaning. A planner therefore cannot request a specific variant of a neighbor's behavior, e.g., a faster or more aggressive one, when building contingency plans or probing worst cases. We propose PrefCVAE, a training scheme that ties one latent dimension of a generative predictor to a chosen behavioral attribute (such as speed), so increasing the latent value monotonically increases that attribute in the predicted trajectories. During training, trajectory pairs are generated from ordered latent values, an oracle ranks each pair by a chosen attribute, and a preference loss aligns the attribute ordering with the latent ordering, acting only on the model's samples while adding no inference cost. We instantiate PrefCVAE on a CVAE predictor with a differentiable average-speed oracle. In converged runs, predicted speed increases monotonically with the assigned latent value for nearly all test agents, accuracy stays comparable to the base model, and the posterior encoder recovers the attribute from generated trajectories. Our results show preference supervision is a promising way to make sampling-based generative predictors controllable.
\end{abstract}


\section{Introduction}
\label{sec:introduction}

Trajectory prediction, a key task for safe autonomous driving, forecasts the behaviors of road participants based on recent motions, accounting for complex and multimodal interactions among road agents \cite{shi2024mtr++, westerhout2023smooth}. Deep generative models represent this uncertainty by sampling, and many families do so with high accuracy. These include conditional variational autoencoders (CVAEs)~\cite{yuan2021agentformer}, generative adversarial networks~\cite{huang2020diversitygan}, normalizing flows~\cite{meszaros2024trajflow}, diffusion models~\cite{li2024bcdiff, huang2026versatile}, flow matching~\cite{fu2025moflow}, and autoregressive token models~\cite{seff2023motionlm, wu2024smart}. Conditional VAEs (CVAEs), in particular, excel in modeling the relationships among future trajectories, historical observations, and latent generative factors \cite{kingma2014semi}.

	\begin{figure}[t]
		\centering
		\includegraphics[width=\linewidth]{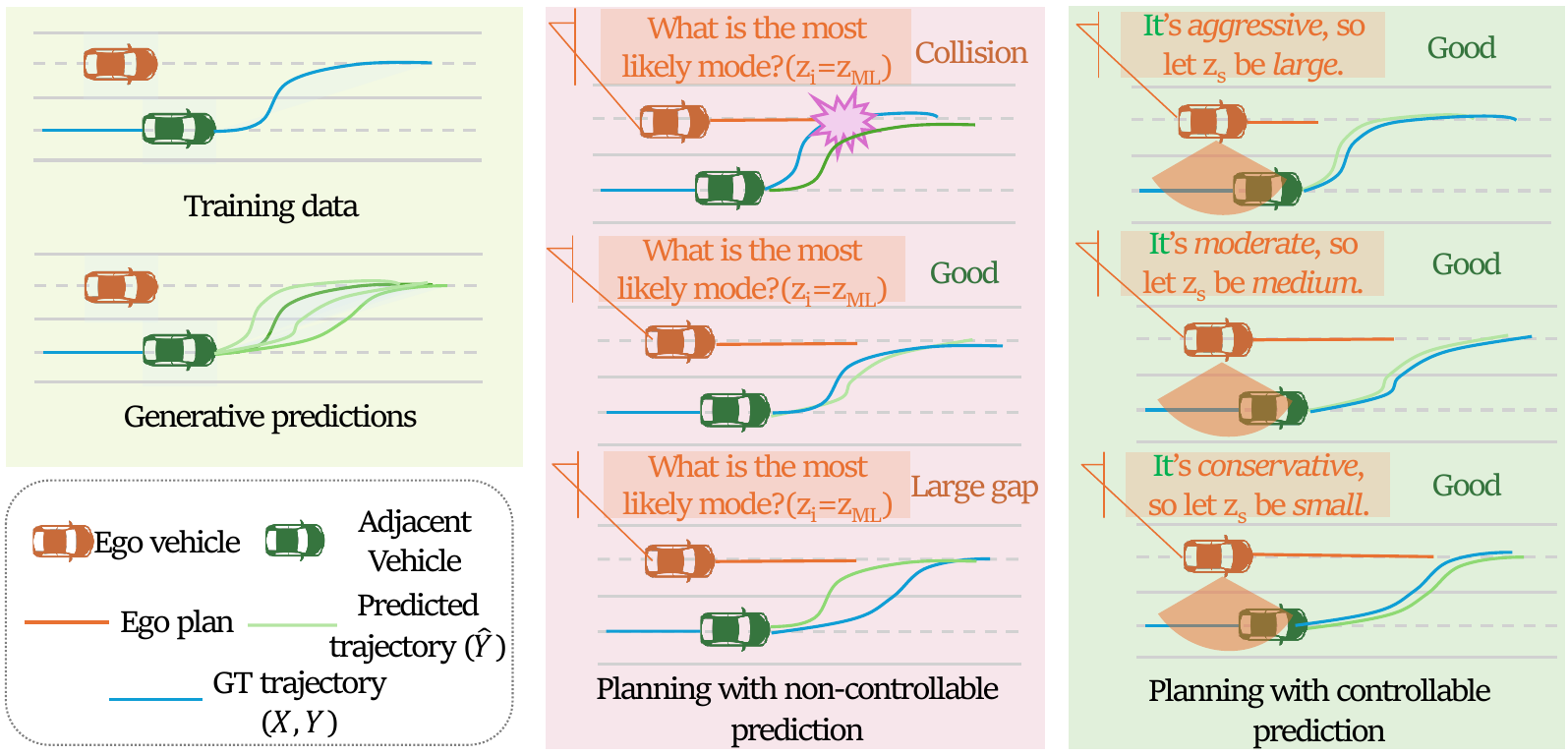}
		\caption{Motivation for controllable prediction. The prediction that is most likely across population is not always the one that materializes. A predictor whose latent variable controls an interpretable attribute (illustrated as driving style; our experiments control average speed, a coarse proxy) lets the planner reason about each variant explicitly.}
		\label{fig:motiv}
	\end{figure}

\begin{figure}[t]
    \centering
    \includegraphics[width=0.92\linewidth]{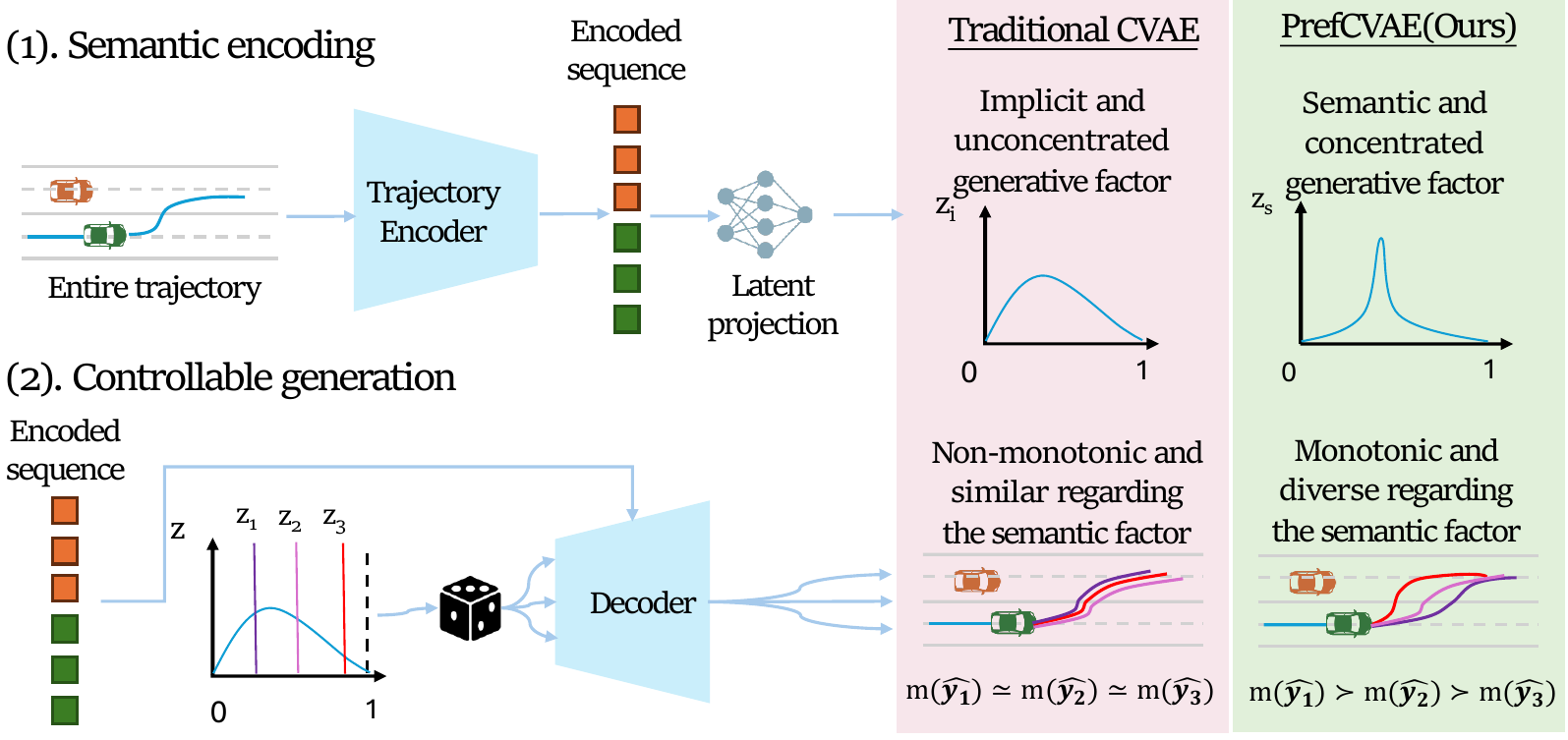}
    \caption{The two capabilities PrefCVAE targets. (1) Semantic encoding: the posterior encoder maps a trajectory to a concentrated latent that orders it by a predefined attribute $m(\cdot)$ given the context. (2) Controllable generation: assigning latent values produces predictions whose attribute is ordered accordingly (line colors match the latent samples on the left; $\succ$: is preferred to, i.e., has the smaller metric).}
    \label{fig:desiderata}
\end{figure}

However, prediction ultimately serves planning~\cite{ivanovic2022injecting, hagedorn2025integration}, and accuracy metrics capture only part of what a planner needs from a predictor~\cite{phong2023truly}. Contingency planning, risk-aware planning, and scenario-based testing reason about \emph{specific} hypotheses, such as ``what if this vehicle merges more aggressively than usual?''~\cite{mustafa2025racp, nishimura2023rap}. Without behavior labels or test-time guidance, generative predictors cannot answer such queries. They are trained to reproduce the distribution of recorded futures given the past~\cite{yuan2021agentformer}, so their samples concentrate on the most likely futures, i.e., the behaviors most common in similar situations in the training data. Moreover, these samples depend on implicit latent variables with no semantic meaning~\cite{paige2017learning}. Therefore, the planner can neither ask for a particular level of a behavioral attribute nor rely on i.i.d.\ sampling to cover the conservative and aggressive ends of the plausible range. A planner that commits to the most likely mode can be caught out by a less likely but plausible behavior. If however predictions could be indexed by an interpretable attribute, this could let the planner consider conservative, moderate, and aggressive variants on purpose (Fig.~\ref{fig:motiv}).

Several existing approaches make predictions controllable. Some rely on behavior labels, computed by rules or annotated, on personas, or on hand-designed reward channels~\cite{huang2020diversitygan, jiao2022tae, lu2026controllable}. Others condition on ego plans or a risk level~\cite{huang2022tip, nishimura2023rap}. A third group conditions or steers diffusion models with scenario tokens, differentiable costs, rules, or language~\cite{pronovost2023scenario, zhong2023guided, zhong2023language}; test-time guidance adds inference cost and requires engineered objectives, but it can change the objective without retraining. However, none of the existing approaches can learn a \emph{continuous}, \emph{monotonic} control of a chosen attribute at no inference cost. 

To address this limitation, we introduce Preference CVAE (PrefCVAE), a simple training scheme for latent-variable trajectory predictors. During training, PrefCVAE decodes two trajectories from two ordered latent values. A fixed, differentiable comparison oracle judges which of the two exhibits more of an attribute, and a preference loss rewards the model when the attribute ordering agrees with the latent ordering, and penalizes if it disagrees. After training, the decoder produces controllable predictions from an assigned latent value, and the posterior encoder recovers this value from generated trajectories (Fig.~\ref{fig:desiderata}). Our contributions are:
\begin{itemize}
	\item \textbf{PrefCVAE: a training method that gives generative predictors monotonic, per-agent control over a chosen behavioral attribute.} Because the supervision acts on the model's own samples, no recorded trajectory needs an attribute or latent label. The resulting preference loss reduces to a weighted pairwise ranking loss backpropagated through a frozen oracle, which makes it simple to add to existing latent-variable predictors.
	\item \textbf{Empirical evidence that PrefCVAE can achieve monotonic, per-agent control with only a minor loss of accuracy.} On nuScenes with a CVAE-based AgentFormer~\cite{yuan2021agentformer}, the best PrefCVAE model reduces violations of speed, a simple case of behavioral attribute, monotonicity from 82\% to 0.1\% of test agents, while $\minADE$ and $\minFDE$ stay within 0.04 and 0.14~m of the base model; at the base model's latent dimension, PrefCVAE matches its accuracy (2.63/5.57~m vs.\ 2.62/5.62~m). The posterior encoder learns the same semantic axis and recovers the assigned latent value of generated trajectories.
	\item \textbf{An empirical rule for sparse preferences.} When only a fraction $\nu$ of the pairs receives a preference, e.g., because queries are costly, raising the loss weight $\lambda$ compensates: control emerges once $\lambda\nu\geq1.6$ (best of three seeds), so with $\lambda=32$, preferences on 5\% of the pairs suffice.
\end{itemize}
Code is available at \url{https://anonymous.4open.science/r/prefcvae-iv27submission}.

\section{Related Work}
\label{sec:related}

\subsection{Generative Trajectory Prediction}
CVAE-based predictors such as AgentFormer~\cite{yuan2021agentformer} model the future as a decoder conditioned on the observed history and a latent variable. Query-based transformers~\cite{shi2024mtr++} and next-token prediction models~\cite{seff2023motionlm, wu2024smart} now set the state of the art. Diffusion and flow-matching models have been applied to prediction~\cite{li2024bcdiff, fu2025moflow} and multi-agent behavior simulation~\cite{huang2026versatile}. These models are optimized for likelihood or accuracy. Their latent or noise variables carry no prescribed meaning, which is the gap this work addresses.

\subsection{Controllable Prediction and Scenario Generation}
\emph{Semantic latents for prediction.} Interpretable latents have been built with expert decoder priors~\cite{neumeier2021variational} and by metric learning on rule-based maneuver labels~\cite{huang2020diversitygan}. TAE~\cite{jiao2022tae} regresses a continuous aggressiveness latent onto time-headway values available for part of the recorded data. PersonaDrive~\cite{lee2026personadrive} conditions prediction on nine annotated personas and enforces their ordering along two discrete axes. SCBG~\cite{chang2023editing} auto-labels recorded trajectories with a continuous courtesy level inspired by social value orientation~\cite{schwarting2019social} and trains a conditional generator through a differentiable courtesy measure of its own outputs; unlike PrefCVAE, it needs absolute attribute values for the data. Concurrently, CNeVA~\cite{lu2026controllable} obtains monotone speed control in a rectified-flow simulator from a latent derived from absolute returns of hand-designed reward channels, sampled with classifier-free guidance. RAP~\cite{nishimura2023rap} adds an encoder conditioned on a continuous risk level to a frozen CVAE decoder, trained from a differentiable planner cost; it biases the forecast toward costly outcomes rather than ordering a latent by a trajectory attribute. 

\emph{Guided generation for simulation and testing.} Diffusion models condition scenario generation on tokens~\cite{pronovost2023scenario} or steer samples toward rules, costs, or language at inference time~\cite{zhong2023guided, zhong2023language, huang2026versatile}. Related methods generate safety-critical scenarios~\cite{chang2024safesim, peng2025safety}.

\emph{Attribute control in generative models.} AR-VAE~\cite{pati2021attribute} makes one VAE latent monotonic in an attribute with a pairwise ordering regularizer, and ROVAE~\cite{chen2019robust} learns an ordinal latent factor from noisy pairwise comparisons. Both regularize the encoder codes of \emph{recorded} samples. PrefCVAE instead compares pairs of trajectories that it \emph{generates} from sampled latents, so the gradient reaches the generator through the oracle and the sampled latent order is the target. Training a generator through a frozen differentiable critic on its own samples has precedents in reward fine-tuning~\cite{clark2024directly} and traffic simulation~\cite{chang2023editing}, and with jointly learned discriminators in controllable text generation~\cite{hu2017toward}. Unlike guided diffusion, PrefCVAE leaves inference unchanged. In our implementation, the recorded speed additionally selects which pairs are compared (Sec.~\ref{sec:objective}).

\subsection{Learning from Preferences and Weak Supervision}
Preference learning fits a reward to pairwise human judgments~\cite{christiano2017deep}, and DPO~\cite{rafailov2024direct} removes the explicit reward model. In driving, preference optimization and RL fine-tuning have improved the scene consistency of predictors~\cite{azevedo2025improving}, the realism of motion generators~\cite{tian2025direct}, and diffusion planners and personalized driving styles~\cite{huang2025gendrive, li2025learning}. These methods align the \emph{whole} output distribution with a preferred behavior, whereas PrefCVAE builds an ordered latent axis. Diffusion models are fine-tuned through differentiable rewards~\cite{clark2024directly} or directly from preference pairs~\cite{wallace2024diffusion}. Weakly supervised disentanglement learns factors from factor labels or from pairs of \emph{recorded} observations that share or are ranked by factors~\cite{locatello2020weakly, shu2020weakly, chen2019robust}. PrefCVAE needs neither: the order of two \emph{sampled} latents is the target, and the gradient flows through a frozen, differentiable oracle that plays the role of the reward model in reward fine-tuning (Sec.~\ref{sec:pref-loss}).

\section{Preliminaries}
\label{sec:prelim}

\subsection{Problem Formulation}
A scene contains $N$ agents. The observed history $\vx = \{\vx_n^t\}$ consists of the states $\vx_n^t$ (e.g., position, velocity, heading) of each agent $n$ at time steps $t \in \{1,\dots,T_\text{cur}\}$, together with map context. The future $\vy = \{\vy_n^t\}$ covers $t \in \{T_\text{cur}+1,\dots,T_\text{end}\}$. Given a dataset $\mathcal{D}$ of $K$ scene samples $(\vx,\vy)$, a generative predictor learns the conditional distribution $p_\theta(\vy\mid\vx)$ and draws multiple future hypotheses $\hvy \sim p_\theta(\vy\mid\vx)$ from it.

\subsection{CVAE-Based Trajectory Prediction}
\label{sec:cvae-frame}
A CVAE introduces an $M$-dimensional latent variable $\vz_n$ for each agent, with $\vz = [\vz_1,\dots,\vz_N]$. It jointly learns a prior $p_\theta(\vz\mid\vx)$, a decoder $p_\theta(\vy\mid\vx,\vz)$, and an approximate posterior (encoder) $q_\phi(\vz\mid\vx,\vy)$. Training minimizes the negative evidence lower bound,
\begin{multline}
    \mathcal{L}_\text{ELBO} = -\mathbb{E}_{q_\phi(\vz\mid\vx,\vy)}\!\left[\log p_\theta(\vy\mid\vx,\vz)\right] \\
    + D_\text{KL}\!\left(q_\phi(\vz\mid\vx,\vy)\,\|\,p_\theta(\vz\mid\vx)\right),
\end{multline}
which AgentFormer~\cite{yuan2021agentformer} complements with a best-of-$k$ variety loss; we denote the sum by $\Lbase$. Nothing in $\Lbase$ ties a latent dimension to a semantic attribute~\cite{locatello2019challenging}.

\section{PrefCVAE}
\label{sec:method}

We split each agent's latent into $P$ semantic dimensions $\vz_{S}$ and $M-P$ implicit dimensions $\vzI$, $\vz_n = [\vz_{S,n}, \vz_{I,n}]$. The semantic dimensions should encode quantifiable attributes. The implicit dimensions capture all remaining variation. In this work we describe and evaluate a single semantic dimension $\zS$. The attribute is given by a scalar metric $m(\vx_n,\vy_n)$, e.g., the average speed of agent $n$. PrefCVAE thus applies to attributes that order trajectories given the context, i.e., for which ``more'' and ``less'' are meaningful. Monotonicity makes the direction in which a change of $\zS$ moves the attribute known, so a planner can search over the values of $\zS$. Without loss of generality, we let smaller latent values correspond to smaller metric values.

\paragraph{Why preferences} PrefCVAE targets attributes for which comparison is meaningful and easy to acquire while absolute scales are ill-defined, such as the aggressiveness of two maneuvers: It is easy for a judge to tell which is more aggressive but quantification is intractable~\cite{christiano2017deep}. For the average speed we studied in this paper, we use it because it is easy to be calculated from an oracle program and allows for exact evaluation of controllability. But even for average speed, it is hard to obtain a good scaling and the preference loss enables the latent to learn it implicitly from the data. Note that human judgments are not differentiable; Sec.~\ref{sec:discussion} describes how they can enter PrefCVAE.


\subsection{Weak Preferences over Generated Trajectories}
\label{sec:weak-pref}
Following the weak-supervision literature, in particular rank pairing~\cite{shu2020weakly}, we call the supervision \emph{weak}: it concerns which of two trajectories exhibits more of the attribute, not the attribute value of either trajectory or a latent value. In our experiments, the comparison is computed by an exact, differentiable metric, which conveys magnitudes and gradients and is therefore more informative than a human label.

\paragraph{Auxiliary latent pairs} For every agent, we draw two latent vectors $\vz^0_n, \vz^1_n \sim \mathcal{U}(0,1)^M$ and order them so that $\zS^0 < \zS^1$ (agent index omitted). The implicit dimensions of the two vectors are sampled independently and are not ordered.

\paragraph{Auxiliary predictions} The decoder generates the mean predictions $\hvy^i = \mathbb{E}[p_\theta(\vy\mid\vx,\vz^i)]$ for $i\in\{0,1\}$. These are generated trajectories decoded from known, sampled latents, so no latent labels are needed.

\paragraph{Oracle preference} A comparison oracle evaluates the metric on both predictions. With a differentiable metric, we define the soft preference
	\begin{equation}
		\Phat = \frac{(\zS^1-\zS^0)\,\sigma\!\big(\eta\,[m(\vx,\hvy^{0})-m(\vx,\hvy^{1})]\big) + \zS^0}{\zS^0+\zS^1},
		\label{eq:pref}
	\end{equation}
where $\sigma$ is the logistic function and $\eta$ sets how sensitive the preference is to the metric difference. $\Phat$ is a smooth \emph{if--else}. It approaches $\zS^1/(\zS^0+\zS^1)$ when $\hvy^0$ has the larger metric (the pair contradicts the latent order; the $\sigma$ term approaches 1) and $\zS^0/(\zS^0+\zS^1)$ when the pair agrees with it (the $\sigma$ term approaches 0). The sigmoid replaces a hard comparison so that the preference stays differentiable with respect to the generated trajectories. Non-differentiable, e.g., human-provided, preferences are discussed in Sec.~\ref{sec:discussion}.

\subsection{Preference Loss}
\label{sec:pref-loss}
Inspired by preference learning~\cite{christiano2017deep}, we write the preference loss as a cross-entropy between the oracle preference and the latent-implied distribution $(\zS^0,\zS^1)/(\zS^0+\zS^1)$:
	\begin{equation}
		\Lpref = -\Big[\Phat\log\frac{\zS^0}{\zS^0+\zS^1} + (1-\Phat)\log\frac{\zS^1}{\zS^0+\zS^1}\Big].
		\label{eq:loss}
	\end{equation}
With $\zS^0<\zS^1$, $\Lpref$ is large when the generated pair contradicts the latent order ($\Phat$ large) and small when it agrees. Swapping the roles of $\zS^0$ and $\zS^1$ reverses the direction of the alignment.

\paragraph{Gradient path} The latents are \emph{sampled}, not inferred, so the latent-implied distribution is a fixed target, and all gradients of~\eqref{eq:loss} flow through $\Phat$ into the decoder and the shared context encoders. This reverses the roles in reward learning from preferences~\cite{christiano2017deep}, where the preference is the fixed label: here, the latent order is the label and the generated pair is trained to agree with it. The oracle itself is not trained; only its derivative $\partial m/\partial\hvy$ tells the generator how to change its outputs. Because~\eqref{eq:loss} is linear in $\Phat$, substituting~\eqref{eq:pref} yields the equivalent objective
	\begin{equation}
		\Lpref = w(\zS^0,\zS^1)\;\sigma\!\big(\eta\,[m(\vx,\hvy^0)-m(\vx,\hvy^1)]\big) + \text{const},
		\label{eq:ranking}
	\end{equation}
with $w = \frac{r-1}{r+1}\log r \geq 0$ and $r=\zS^1/\zS^0$. Although~\eqref{eq:loss} has the form of a cross-entropy, it is linear in $\Phat$ and is minimized at a boundary rather than by matching two distributions. It is a weighted sigmoid (smoothed 0--1) ranking loss, and minimizing it makes the prediction with the smaller latent value exhibit the smaller metric value. The weight depends only on the latent ratio: it diverges as $\zS^0\to0$ and is small in the upper part of the latent range, so pairs whose smaller latent is close to zero dominate. With our $\eta$ (Sec.~\ref{sec:exp-setup}), the sigmoid saturates only for speed differences above about 6~m/s. Typical training pairs stay below that, so the loss keeps pushing the attribute apart and $\Lbase$ bounds the spread; only widely separated pairs of fast agents saturate (Fig.~\ref{fig:res1-vanilla-af}). Unlike the logistic ranking loss $-\log\sigma(\cdot)$, the sigmoid also flattens for strongly misordered pairs. Because the loss is evaluated on generated rather than recorded trajectories, it does not confine the latent range to the attribute range observed in the data.

	\subsection{Training Objective and Use Rate}
	\label{sec:objective}
	We train with
	\begin{equation}
		\mathcal{L} = \Lbase + \frac{\lambda}{|\mathcal{K}|}\sum_{n\in\mathcal{K}} b_n\,\mathcal{L}_{\text{pref},n}, \qquad b_n\sim\text{Bernoulli}(\nu),
		\label{eq:total}
	\end{equation} 
where $\lambda$ weights the preference loss and $\mathcal{K}$ is the set of agents whose auxiliary pair is evaluated. The \emph{use rate} $\nu$ is the probability that a generated pair is sent to the oracle. The decision $b_n$ is drawn independently per agent and iteration. A pair without a preference ($b_n=0$, i.e., $\Phat=\tfrac{1}{2}$) contributes no gradient. Because such pairs remain in the normalizer $|\mathcal{K}|$, the expected preference term scales with $\lambda\nu$, so $\nu$ sets both the sparsity and the effective weight of the preference signal. In our implementation, $\mathcal{K}$ contains only \emph{significant} comparisons: pairs in which both predictions' average speeds deviate by more than 30\% from the ground-truth future speed of the agent. We use this filter in all experiments. Clearly deviating pairs provide a strong preference signal, which makes the ordering easier to learn, while predictions close to the recorded future are left to the reconstruction term. 

Note that each semantic dimension receives its own oracle and preference loss. Sampling the pair so that it differs only in that dimension isolates its effect on the decoder. Several dimensions can therefore be controlled independently when their attributes are compatible, e.g., speed and lateral offset. Hence, generalizing to multiple semantic dimensions is technically straightforward. See Sec.~\ref{sec:discussion} for details.

\paragraph{Bounded latent} Instead of the usual Gaussian latent, we use a Beta distribution, whose support is $[0,1]$. The attribute values that occur in traffic are bounded in practice even if the metric itself is not; a bounded latent with known endpoints makes it clear which values to assign at test time.

	\section{Experiments}
	\label{sec:experiments}

	\begin{figure*}[t]
		\centering
		\subfloat[]{\includegraphics[width=0.74\textwidth]{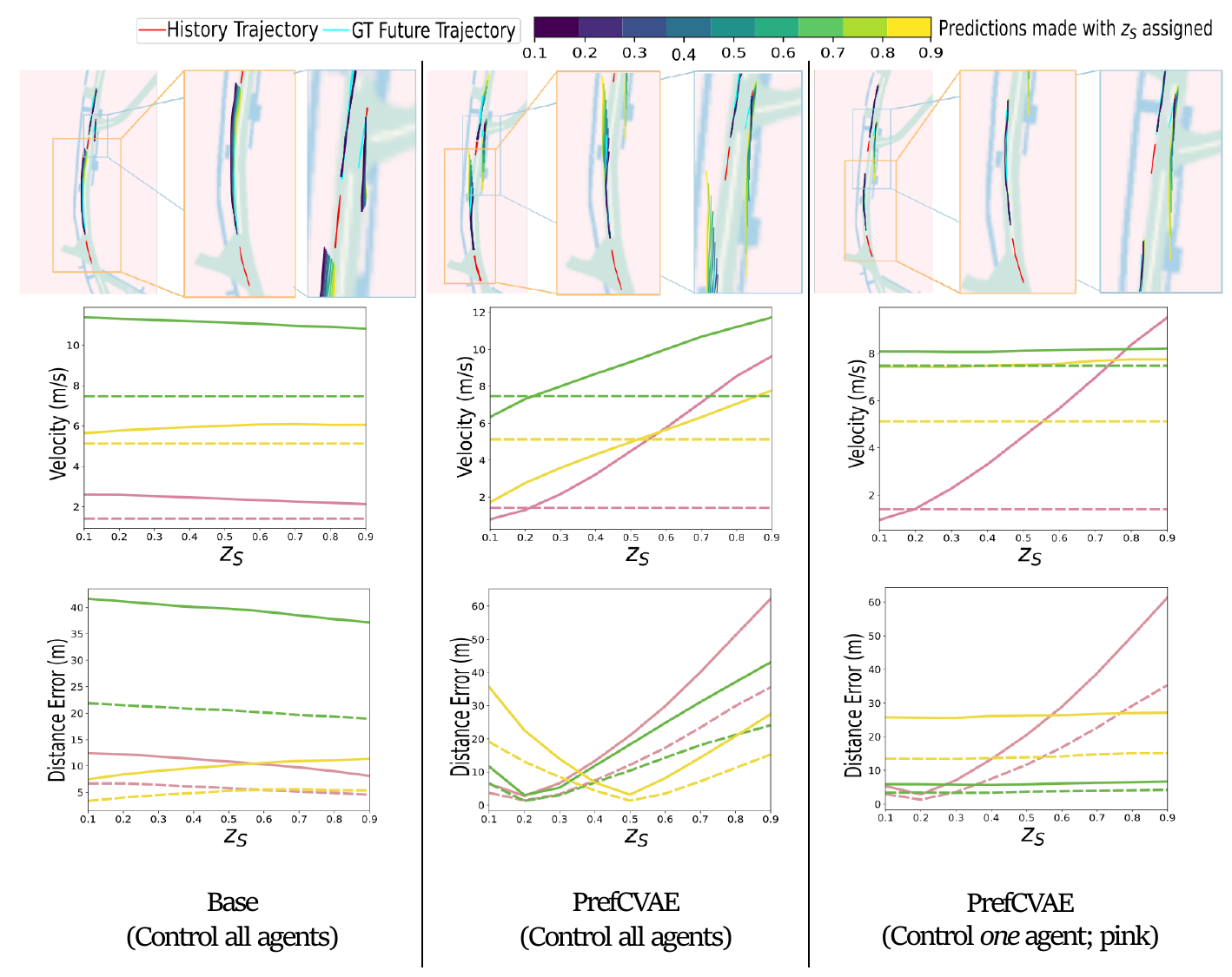}\label{fig:res1-vanilla-af}}
		\hfill
		\subfloat[]{\includegraphics[width=0.2\textwidth]{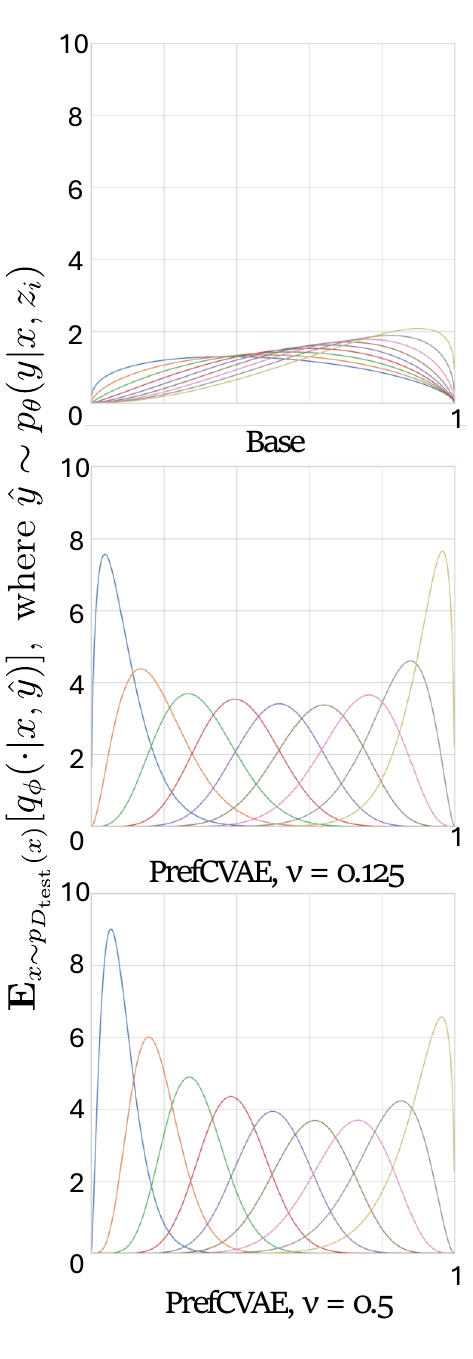}\label{fig:regressed}}
		\caption{(a) Controllable predictions from the PrefCVAE decoder on a test scene. Columns: base model, PrefCVAE controlling all agents, and PrefCVAE controlling only one agent (pink) while the latents of the others are drawn once and held fixed. Rows: predictions colored by the assigned $\zS$; average speed versus $\zS$ (dashed lines: ground truth); displacement error of the first sample versus $\zS$ (solid: $\sqrt{12}\times$ the root-mean-square error over the 12 predicted steps; dashed: final displacement error). With PrefCVAE ($\nu=0.125$), a larger $\zS$ yields a higher average speed, and the error is lowest near the latent value whose speed matches the ground truth. (b) Posterior distributions of $\zS$ encoded from test-set predictions generated with nine assigned values $\zS\in\{0.1,\dots,0.9\}$ (one color each). Ideally they are nine narrow peaks at the assigned values.}
		\label{fig:res1}
	\end{figure*}

	\subsection{Experimental Setup}
	\label{sec:exp-setup}

\paragraph{Dataset and base model} We use the nuScenes~\cite{caesar2020nuscenes} prediction benchmark and predict 6~s (12 frames) from 1.5~s (4 frames) of history, following the data split of AgentFormer~\cite{yuan2021agentformer}. The base model, $\beta$-AgentFormer, is AgentFormer with its Gaussian latent replaced by a Beta latent. The concentration parameters are $\alpha,\beta = \text{ELU}(\cdot)+2 > 1$, which keeps the distribution unimodal with its mode in $(0,1)$. We exclude AgentFormer's DLow trajectory sampler~\cite{yuan2020dlow} so that the latent space is evaluated directly. For reference, AgentFormer with DLow reports 1.86/3.89~m $\minADE$/$\minFDE$~\cite{yuan2021agentformer}; our claims concern changes relative to $\beta$-AgentFormer. Without DLow, the five samples are i.i.d.\ draws from the prior. All models are trained from scratch with $\Lbase$, or with $\Lbase$ plus the preference loss, on the full training set for 50 epochs with Adam. The base model is $\beta$-AgentFormer with $M=32$ and a learned Beta prior $p_\theta(\vz\mid\vx)$, trained with $\Lbase$ only. PrefCVAE uses a fixed Beta(1,1) prior and, unless stated otherwise, $M=2$, $\lambda=16$, and the first latent dimension as $\zS$. The oracle metric $m$ is the mean per-step speed over the 12 predicted steps. We use $\eta=10$ on AgentFormer's normalized coordinates, which corresponds to $\sigma(0.5\,\Delta v)$ for a speed difference $\Delta v$ in m/s.

\paragraph{Evaluation Metrics} \label{exp:metrics}
Accuracy is not the main object of this work, but we report $\minADE$ and $\minFDE$ to verify that controllability does not cost accuracy. These metrics are computed by sampling the latent from the prior, without assigning it. To evaluate the decoder, we assign $\zS\in\{0.1,0.2,\dots,0.9\}$ to all agents and measure the resulting average speed. The \emph{speed range} reports the minimum and maximum of the test-set mean speed over the nine assigned values. The \emph{violation rate} (VR) measures how consistently the control is monotonic. A violation occurs when $\zS^a>\zS^b$ but the prediction for $\zS^a$ is slower than the prediction for $\zS^b$. VR is evaluated on the first of the five samples. The test set contains 138 scenes, 3{,}076 minibatches (one scene at one time step, 4 past and 12 future frames), and 9{,}041 agents. VR is reported agent-wise (AW: fraction of violating agents), minibatch-wise (MBW: fraction of minibatches with at least one violation), and scene-wise (SW: fraction of scenes with at least one violation). To evaluate the encoder, we encode each of the nine prediction sets with $q_\phi$ and fit a Beta distribution $\hat q_j$ to the encoded latents of set $j$. We then report the average pairwise Jensen--Shannon divergence (JSD) between the nine fits (distinctness), the cumulative log-likelihood $\sum_{j=1}^{9}\log \hat q_j(\zS=\tfrac{j}{10})$ at the assigned values (concentration), and the mean absolute deviation of each fit's mean $\hat\alpha_j/(\hat\alpha_j+\hat\beta_j)$ from its assigned value (accuracy).

\paragraph{Repeatability} How well a VAE latent aligns with a factor depends strongly on the random seed~\cite{locatello2019challenging}. We therefore run every configuration with three random seeds and fix the seeds of all other sampling steps. Fig.~\ref{fig:vr-vs-ur} shows the mean and one standard deviation over seeds; Fig.~\ref{fig:res1} shows a single model. Tables report the seed with the lowest agent-wise VR. Importantly, for robustness of the evaluation, we only use the validation set VR to decide if a checkpoint is good or not, so neither test set nor accuracy are involved, and a failed run could be detected on training scenes and restarted. Hence, using the best seed does not imply test set leakage. The gap between converged and failed runs (below 0.5\% vs.\ tens of percent, Table~\ref{tab:ablation2-2}) also far exceeds the VR's sampling noise over 9{,}041 test agents. Tables thus show what converged runs achieve; Fig.~\ref{fig:vr-vs-ur} shows how often runs converge.

	\subsection{Controllable Trajectory Prediction}
	\label{sec:ETP}

	\begin{table}[t]
		\centering
		\caption{Accuracy (latent sampled from the prior), speed range over the nine assigned values of $\zS$, and agent-wise VR of the base model ($M=32$, learned prior) and of PrefCVAE ($M=2$, fixed prior) with different use rates $\nu$ (best of three seeds).}
		\label{tab:general-compare}
		\begin{tabular}{lccc}
			\toprule
			& \multicolumn{3}{c}{Use rate $\nu$} \\
			\cmidrule{2-4}
			& 0 (Base) & 0.125 & 0.5\\
			\midrule
			$\minADE$ (m)     & 2.62      & 2.66      & 2.64      \\
			$\minFDE$ (m)     & 5.62      & 5.76      & 5.74      \\
			\midrule
			Speed min/max (m/s) & 6.32/6.34 & 3.74/8.76 & 3.26/9.72 \\
			AW VR (\%)          & 82.28     & 0.10      & 0.12      \\
			\bottomrule
		\end{tabular}
	\end{table}

	\begin{table}[t]
		\centering
		\caption{Posterior encoding of the assigned values of $\zS$ (base: $M=32$; PrefCVAE: $M=2$). Avg.\ JSD: mean over the 36 pairs of the nine fitted distributions; $\log L$ at the assigned values (both higher is better); deviation of the fitted mean from the assigned value (lower is better).}
		\label{tab:encoding}
		\begin{tabular}{lccc}
			\toprule
			& Base & $\nu$=0.125 & $\nu$=0.5 \\
			\midrule
			Avg.\ JSD                & 0.0352 & \textbf{0.4874}  & 0.4578 \\
			$\log L$                 & 3.19   & \textbf{13.23}   & 11.76  \\
			Avg.\ mean deviation     & 0.1660 & \textbf{0.0090}  & 0.0163 \\
			\bottomrule
		\end{tabular}
	\end{table}

	\paragraph{PrefCVAE can make controllable predictions} Fig.~\ref{fig:res1-vanilla-af} shows a test scene in which the semantic latent is traversed from 0.1 to 0.9. In the base model (left column), we traverse the first of its 32 latent dimensions, which has no consistent relation to speed. Its effect on speed has no consistent sign across agents (Fig.~\ref{fig:res1-vanilla-af}), so the test-set mean speed stays at 6.32--6.34~m/s, and 82\% of the agents violate monotonicity (Table~\ref{tab:general-compare}). With PrefCVAE (middle column), the average speed of the predictions increases monotonically with $\zS$. The test-set mean speed spans 3.74--8.76~m/s at $\nu=0.125$ and 3.26--9.72~m/s at $\nu=0.5$, with $\minADE$/$\minFDE$ within 0.04/0.14~m of the base model. At the base model's $M=32$, PrefCVAE matches its accuracy (2.63/5.57 vs.\ 2.62/5.62~m, Table~\ref{tab:tradeoff}). The displacement error is lowest near the latent value whose speed matches the ground truth, so the recorded future lies within the controlled range. We do not evaluate the realism of the extreme variants. All agents' latents are regularized jointly during training. Even so, in this scene, assigning the latent of a single agent changes the behavior of that agent alone (right column). The latents of the other agents are drawn once and held fixed. This suggests that a planner can query individual agents, although we show it for one scene only.

	\paragraph{PrefCVAE can encode the attribute} We encode the nine sets of predictions generated with $\zS=0.1,\dots,0.9$ back into the latent space, using the assigned values as pseudo ground truth. With PrefCVAE, the encoded distributions line up in the order of their assigned values (Fig.~\ref{fig:regressed}), with an average pairwise JSD of 0.487 and a mean deviation of only 0.009 between each fitted mean and its assigned value (Table~\ref{tab:encoding}). For the base model, whose latent dimensions carry no designated meaning, the nine distributions nearly coincide (JSD 0.035). The posterior encoder receives no gradient from $\Lpref$ (Sec.~\ref{sec:pref-loss}), so this alignment emerges through the ELBO: the encoder learns to invert a decoder that has become monotonic in $\zS$. This is a cycle-consistency test on generated trajectories. If the property carries over to recorded trajectories, which we have not tested, the encoder could annotate data with values of $\zS$ (Sec.~\ref{sec:discussion}).

	\subsection{Hyperparameter Analysis}
	\label{sec:ablation}
	
	\begin{figure}[t]
		\centering
		\includegraphics[width=0.85\linewidth]{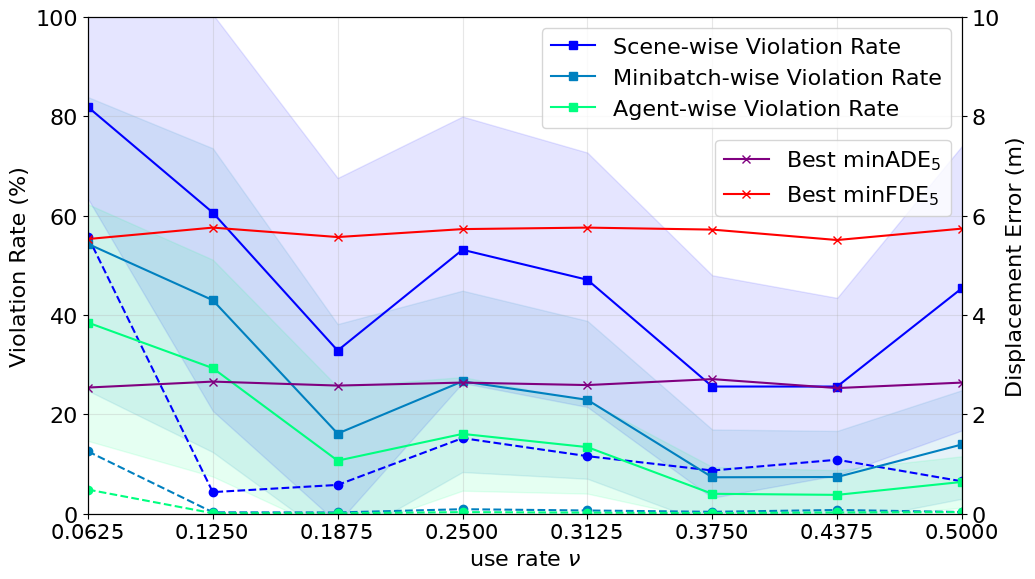}
		\caption{Violation rates (solid: mean over three seeds; shaded: one standard deviation; dashed: best seed) and best-seed $\minADE$/$\minFDE$ (right axis) versus use rate $\nu$ ($\lambda=16$, $M=2$).}
		\label{fig:vr-vs-ur}
	\end{figure}

	\paragraph{Use rate} To test how sparse the preference signal can be, we randomly drop preferences. For every $\nu\geq0.125$, the best of three seeds reaches an agent-wise VR below 0.5\% (Fig.~\ref{fig:vr-vs-ur}, dashed). The differences between these best seeds are at most about 20 of 9{,}041 agents (0.10--0.32\% AW VR). Averaged over seeds (solid), however, the VR is high at every use rate. It generally decreases with $\nu$ and is lowest at $\nu\in\{0.375,0.4375\}$. Not every seed converges to a monotonic latent, and convergence tends to become less reliable as preferences become sparser. By~\eqref{eq:total}, lowering $\nu$ also lowers the effective weight $\lambda\nu$, which we examine next.

	\begin{table}[t]
		\centering
		\caption{Violation rates (\%, best of three seeds) for weights $\lambda=8$ and $32$ at small use rates.}
		\label{tab:ablation2-2}
		\setlength{\tabcolsep}{4pt}
		\begin{tabular}{lcccccc}
			\toprule
			& \multicolumn{2}{c}{SW VR} & \multicolumn{2}{c}{MBW VR} & \multicolumn{2}{c}{AW VR} \\
			\cmidrule(lr){2-3} \cmidrule(lr){4-5} \cmidrule(lr){6-7}
			$\nu$ & $\lambda$=8 & $\lambda$=32 & $\lambda$=8 & $\lambda$=32 & $\lambda$=8 & $\lambda$=32 \\
			\midrule
			0.025 & 91.30 & 96.38 & 64.56 & 64.11 & 45.99 & 38.70\\
			0.050 & 93.48 & 10.14 & 65.08 & 0.62 & 46.09 & 0.21 \\
			0.075 & 91.30 & 13.04 & 63.30 & 1.07 & 44.62 & 0.38 \\
			0.100 & 95.65 & 11.59 & 77.44 & 0.94 & 56.89 & 0.32 \\
			0.125 & 38.41 & 15.22 & 6.05  & 0.94 & 2.18 & 0.32 \\
			\bottomrule
		\end{tabular}
	\end{table}

	\paragraph{Weighting factor} The experiments above use $\lambda=16$. We also train with $\lambda=8$ and $32$ for $\nu\in[0.025,0.125]$, the regime where the violation rate is not yet saturated. With $\lambda=8$, the best seed fails for $\nu\leq0.10$. With $\lambda=32$, it reaches a low VR for every $\nu\geq0.05$ (Table~\ref{tab:ablation2-2}). At $\nu=0.125$, $\lambda=16$ is better than both. Across Table~\ref{tab:ablation2-2} and Fig.~\ref{fig:vr-vs-ur}, the best seed fails for $\lambda\nu\leq0.8$, is partially monotonic at $\lambda\nu=1$ ($\lambda=8,\nu=0.125$ and $\lambda=16,\nu=0.0625$), and reaches a low VR for $\lambda\nu\geq1.6$. Best-seed success is thus consistent with a threshold on $\lambda\nu$, and with $\lambda=32$ the best seed reaches a low VR with preferences on 5\% of the pairs. How reliably seeds converge does not follow $\lambda\nu$ alone (Fig.~\ref{fig:vr-vs-ur}, solid). Accuracy changes little ($\minADE$ 2.51--2.73~m, $\minFDE$ 5.46--5.97~m).

	\begin{table}[t]
		\centering
		\caption{Effect of the latent dimension $M$ on violation rate, speed range, and accuracy (best of three seeds, $\nu=0.125$).}
		\label{tab:tradeoff}
		\setlength{\tabcolsep}{4pt}
		\begin{tabular}{lcccccc}
			\toprule
			& \multicolumn{2}{c}{$M=8$} & \multicolumn{2}{c}{$M=16$} & \multicolumn{2}{c}{$M=32$} \\
			\cmidrule(lr){2-3} \cmidrule(lr){4-5} \cmidrule(lr){6-7}
			& $\lambda$=16 & $\lambda$=32 & $\lambda$=16 & $\lambda$=32 & $\lambda$=16 & $\lambda$=32 \\
			\midrule
			SW VR (\%)   & 3.62 & 0.72 & 5.07 & \textbf{0} & 0.72 & 5.80 \\
			MBW VR (\%)  & 0.20 & 0.03 & 0.23 & \textbf{0} & 0.03 & 0.26 \\
			AW VR (\%)   & 0.07 & 0.01 & 0.08 & \textbf{0} & 0.01 & 0.09 \\
			\midrule
			Speed min (m/s) & 3.48 & 3.10 & 3.48 & 3.44 & 3.60 & 3.36  \\
			Speed max (m/s) & 10.06 & 12.00 & 10.16 & 11.54 & 9.92 & 9.82 \\
			\midrule
			$\minADE$ (m) & 2.70 & 2.83 & 2.70 & 2.83 & \textbf{2.63} & 2.66 \\
			$\minFDE$ (m) & 5.75 & 5.91 & 5.73 & 5.91 & \textbf{5.57} & 5.61 \\
			\bottomrule
		\end{tabular}
	\end{table}

	\paragraph{Latent dimension} We test $M\in\{8,16,32\}$ (Table~\ref{tab:tradeoff}). Among these, $M=32$ gives the best accuracy. With $\lambda=32$, the speed range narrows from 8.90~m/s ($M=8$) and 8.10~m/s ($M=16$) to 6.46~m/s ($M=32$). With $\lambda=16$, it is nearly unchanged (6.32--6.68~m/s), while $M=2$ at the same $\nu=0.125$ (Table~\ref{tab:general-compare}) spans 5.02~m/s. The effect of $M$ on controllability is therefore not monotonic, and we do not attribute it to a single mechanism. Unsupervised disentanglement~\cite{chen2018isolating} might widen the controllable range at large $M$.



\section{Discussion}
\label{sec:discussion}

\paragraph{Controllable predictions for planning} A planner can query, e.g., $\zS\in\{0.1,0.5,0.9\}$ for the agent that matters, with the other agents' latents fixed, instead of hoping that i.i.d.\ samples cover the extremes. This suits contingency planning against a few distinct hypotheses and targeted scenario-based testing~\cite{mustafa2025racp, chang2024safesim}. Because the prior on $\zS$ is uniform and the decoder is monotonic in $\zS$, assigning $\zS=q$ may approximate the $q$-quantile of the predicted speed, which would make these latent values percentile hypotheses. We have not verified this calibration. These values are relative to the context, not absolute speeds. For a given path, average speed roughly sets when an agent reaches a conflict zone, on which gap acceptance and yielding depend, so a value of $\zS$ approximates an earlier- or later-arrival hypothesis. It does not separate yielding from turning or stopping. Fixing the other agents' latents gave an essentially non-reactive what-if in Fig.~\ref{fig:res1-vanilla-af}; whether the joint decoder lets them react in general is untested. A planner could also search for the value of $\zS$ at which its plan becomes unsafe. Closed-loop evaluation is the main next step.

\paragraph{Beyond CVAEs} We use a CVAE because it provides an encoder and a decoder for the same explicit latent, so a value of $\zS$ can be both assigned and recovered, and because sampling needs no iterative denoising or guidance. Diffusion and flow-matching predictors are more expressive and often more accurate, but they sample iteratively and are steered through conditioning or test-time guidance. The preference loss itself only requires samples that are differentiable in the model parameters given a settable latent, plus a comparison oracle. Diffusion and flow-matching predictors~\cite{huang2026versatile, fu2025moflow} meet the first requirement once conditioned on $\zS$. Without an encoder, however, nothing couples $\zS$ to the data, so the preference loss alone would have to shape it. (i) \emph{End to end}: apply~\eqref{eq:ranking} to pairs of samples generated with $\zS^0<\zS^1$, ideally from shared noise, and backpropagate through a truncated sampling chain or a one-step denoised estimate, as in direct reward fine-tuning~\cite{clark2024directly}. (ii) \emph{Two stages}: if the posterior encoder of a pretrained PrefCVAE also recovers values of $\zS$ on recorded trajectories (Table~\ref{tab:encoding} covers generated ones), use it to annotate training trajectories with $\zS$. Then train a diffusion or flow model conditioned on $(\vx,\zS)$, optionally fine-tuned with $\Lpref$. Latent diffusion in a VAE latent space~\cite{peng2025safety} and flows conditioned on behavior latents~\cite{lu2026controllable} already exist, and PrefCVAE could order such latents by a chosen attribute. This is to be studied in future works.

\paragraph{Human preferences} A human label carries no derivative, so the gradient path of Sec.~\ref{sec:pref-loss} is unavailable, and three routes are possible. (i) \emph{Preference model as oracle}: fit a differentiable Bradley--Terry model to human pairs~\cite{christiano2017deep}, freeze it, and use it as $m$, as differentiable reward fine-tuning does~\cite{clark2024directly}. Driving reward models trained on pairwise preferences already exist~\cite{huang2025gendrive}. Our experiments correspond to an ideal preference model, so the use-rate study measures the tolerated sparsity of the preference signal, not the number of human labels. (ii) \emph{Fixed label, gradient on the model}: re-encode the pair, $\tilde\vz^i\sim q_\phi(\vz\mid\vx,\hvy^i)$, and use the human label as the target in~\eqref{eq:loss}. The encoder becomes the preference model. (iii) \emph{Likelihood ratio}: treat the labeled pair as fixed data and apply a Bradley--Terry loss, related to DPO~\cite{rafailov2024direct}, to the decoder log-likelihoods of the preferred versus the swapped assignment of the two trajectories to the two latents. It needs only a (surrogate) likelihood, so it also applies to diffusion models~\cite{wallace2024diffusion}.

\paragraph{Limitations} We studied a single low-level attribute (average speed) computed by an exact, differentiable oracle, on one dataset and one backbone. Average speed is a coarse descriptor of behavior in urban interactions such as yielding or turning. Attributes that matter more for planning, such as aggressiveness or social value orientation~\cite{schwarting2019social}, and human-provided preferences remain to be tested. Several semantic dimensions would interact through the shared decoder and need to be kept disentangled~\cite{chen2018isolating}. Importantly, training is seed-sensitive: averaged over seeds, the VR is high at every use rate (Fig.~\ref{fig:vr-vs-ur}), and only the best seed reaches a low VR at small effective weights. We hypothesize that, as the posterior encoder receives no gradient from $\Lpref$, reconstruction and preference can pull the decoder in different directions; passing the preference signal to the encoder may help. The attribute is fixed at training time, whereas test-time guidance can switch objectives. Finally, our $\beta$-AgentFormer (2.62/5.62~m) is less accurate than AgentFormer with DLow and far from current nuScenes predictors. Whether the preference loss leaves a near state-of-the-art predictor equally accurate and controllable is untested.

\section{Conclusion}
\label{sec:conclusion}
We presented PrefCVAE, a training scheme that aligns a latent dimension of a generative trajectory predictor with a semantic attribute through pairwise preferences over generated trajectories. The preference loss is a weighted sigmoid ranking loss that backpropagates through a differentiable oracle, which adds no inference cost. On nuScenes, it makes a CVAE predictor monotonically controllable in average speed at comparable accuracy in converged runs. For the best of three seeds, control emerges once the loss weight times the fraction of pairs with a preference reaches a threshold, so with a larger weight, preferences on 5\% of the pairs suffice. This could let planners ask for specific, interpretable behavior hypotheses instead of relying on random samples. Future work should include evaluating our approach in closed loop, with non-differentiable human preferences and with other generator families.


\bibliographystyle{IEEEtran}
\bibliography{root}

\end{document}